\documentclass{article}
\usepackage[dvipsnames]{xcolor}
\usepackage{spconf,amsmath,epsfig,xcolor,graphicx,url}
\usepackage{xspace}
\usepackage{booktabs}
\usepackage{multirow}
\usepackage{hyperref}

\definecolor{citecolor}{HTML}{0071BC}
\definecolor{linkcolor}{HTML}{ED1C24}

\hypersetup{colorlinks=True, citecolor=citecolor, urlcolor=magenta}

\newcommand{\dataset}{\textit{Chesapeake Roads Spatial Context (RSC)}\xspace}

\title{Seeing the roads through the trees:\\A benchmark for modeling spatial dependencies with aerial imagery}
\name{\begin{tabular}{c}
     Caleb Robinson\sthanks{Corresponding author: caleb.robinson@microsoft.com}\textsuperscript{1}, Isaac Corley\textsuperscript{2}, Anthony Ortiz\textsuperscript{1}, Rahul Dodhia\textsuperscript{1} \\
     Juan M. Lavista Ferres\textsuperscript{1}, Peyman Najafirad\textsuperscript{2}
\end{tabular}}

\address{Microsoft AI for Good Research Lab\textsuperscript{1}, University of Texas at San Antonio\textsuperscript{2}}

\begin{document}
\maketitle
\begin{abstract}
Fully understanding a complex high-resolution satellite or aerial imagery scene often requires spatial reasoning over a broad relevant context. The human object recognition system is able to understand object in a scene over a long-range relevant context. For example, if a human observes an aerial scene that shows sections of road broken up by tree canopy, then they will be unlikely to conclude that the road has actually been broken up into disjoint pieces by trees and instead think that the canopy of nearby trees is occluding the road. However, there is limited research being conducted to understand long-range context understanding of modern machine learning models.
In this work we propose a road segmentation benchmark dataset, \dataset, for evaluating the spatial long-range context understanding of geospatial machine learning models and show how commonly used semantic segmentation models can fail at this task. For example, we show that a U-Net trained to segment roads from background in aerial imagery achieves an 84\% recall on unoccluded roads, but just 63.5\% recall on roads covered by tree canopy despite being trained to model both the same way. We further analyze how the performance of models changes as the relevant context for a decision (unoccluded roads in our case) varies in distance. We release the code to reproduce our experiments and dataset of imagery and masks to encourage future research in this direction -- \url{https://github.com/isaaccorley/ChesapeakeRSC}.
\end{abstract}
\section{Introduction}
\label{sec:intro}

Deep convolutional neural networks (CNN) and vision transformers (ViT) have shown impressive performance in geospatial machine learning tasks including land use and land cover (LULC) segmentation, scene understanding and classification, and building detection and segmentation~\cite{niu2021hybrid,robinson2022fast,kirillov2023segment}. It has been shown that, unlike the human vision system, modern neural networks are often biased towards local textures and other local features while ignoring long-range dependencies even when global information is available~\cite{malkin2020mining,geirhos2018imagenet,baker2020local}. This phenomenon is often overlooked since models can still perform well in most common benchmark datasets while only using local features. For example, Brendel et al. show that a bag of 32 × 32 features can achieve high performance (87.6\% top-5 accuracy) on ImageNet~\cite{brendel2018approximating}. 

Other vision applications like Visual Question Answering (VQA) require models to perform spatial reasoning~\cite{bendre2021show} and, with the success of general purpose language models, there has been an explosion of research adapting language models to be able to capture long-range dependencies using transformers~\cite{ma2022mega, yu2023megabyte}. Similarly there has been a revival of recurrent neural networks (RNN)~\cite{orvieto2023resurrecting} via state space models (SSM)~\cite{wang2022pretraining, gu2021efficiently, gu2022parameterization, li2022makes, smith2022simplified, mehta2022long, dao2022hungry} to avoid the quadratic cost of attention when modeling long sequences. These methods have recently been successfully applied to modeling images as sequences for image classification~\cite{nguyen2022s4nd} and generation~\cite{peebles2023scalable, yan2023diffusion} as a replacement to their fully convolutional counterparts.

\begin{figure*}[t]
    \centering
    \includegraphics[width=0.9\textwidth]{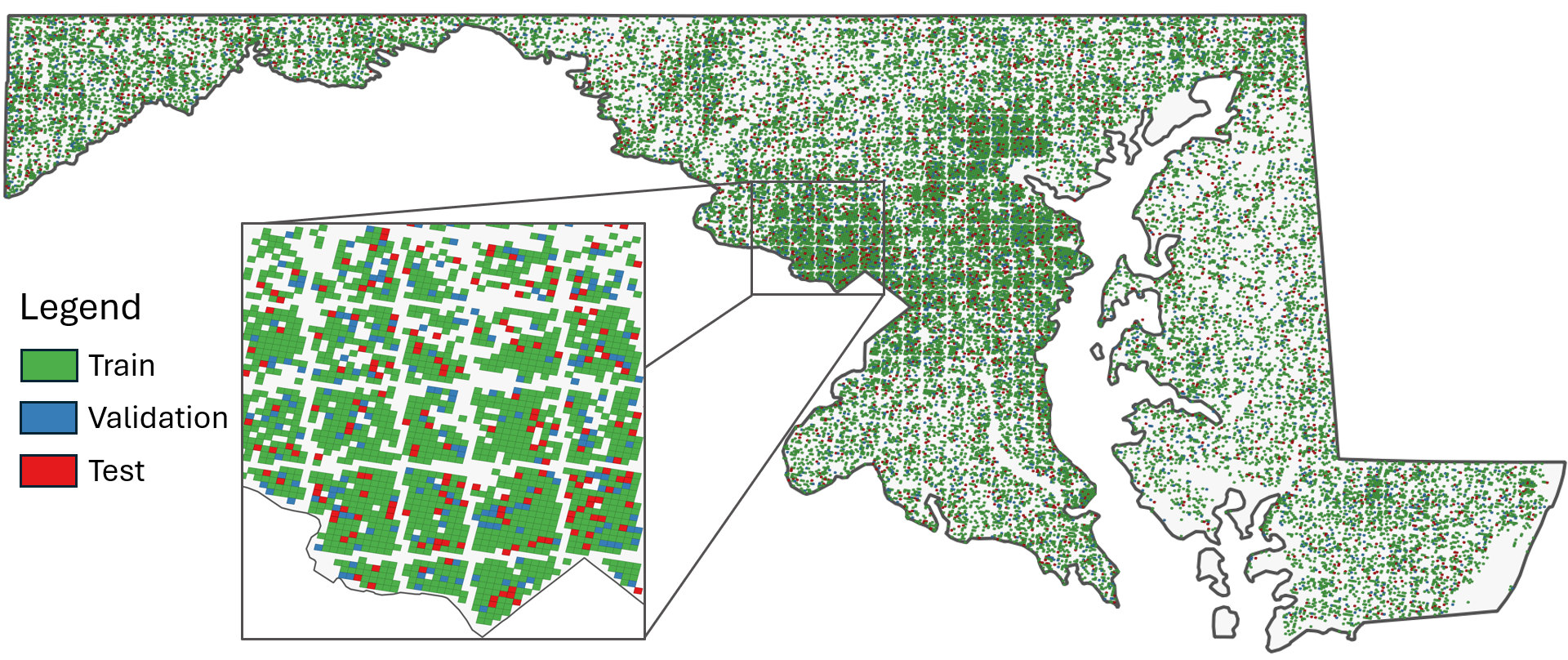}
    \caption{Map of the distribution of the 30,000 train, validation, and test patches in the \dataset dataset.}
    \label{fig:map}
\end{figure*}

Multiple geospatial machine learning applications require models that are able to understand longer range dependencies in imagery. For example, identifying burn scars~\cite{li2004technique}, estimating road network connectivity under occlusions from tree canopy or shadows, and identifying specific land use classes are examples of such applications~\cite{bastani2018roadtracer, bastani2021beyond, bahl2022single}. However, there are no existing benchmark datasets designed specifically to test the long-range spatial reasoning capabilities of existing machine learning models in remote sensing settings. In this work we present a novel semantic segmentation dataset, \dataset, containing high-resolution aerial imagery and labels including ``background'', ``road'' and ``tree canopy over road'' categories which we use to evaluate a machine learning model's ability to incorporate long-range spatial context into predictions. Additionally, we perform an analysis of the long-range reasoning capabilities of multiple canonical segmentation models and find that performance decreases as a function of distance away from the necessary context needed to make a correct prediction. We release our code on GitHub and dataset publicly on HuggingFace\footnote{Links to \href{https://github.com/isaaccorley/ChesapeakeRSC}{code} and \href{https://huggingface.co/datasets/torchgeo/ChesapeakeRSC}{data}.} and through the TorchGeo library~\cite{stewart2022torchgeo}.

To summarize, our contributions are the following:
\begin{itemize}
    \item We propose a new benchmark dataset, \dataset, for measuring the ability of a semantic segmentation model to incorporate long-range context information for the task of segmenting a road network with large gaps.
    \item We benchmark canonical semantic segmentation models on our proposed dataset.
    \item We propose an evaluation workflow that shows how a model's performance varies with distance away from necessary spatial context.
\end{itemize}

\begin{figure*}[t]
    \centering
    \includegraphics[width=0.25\textwidth]{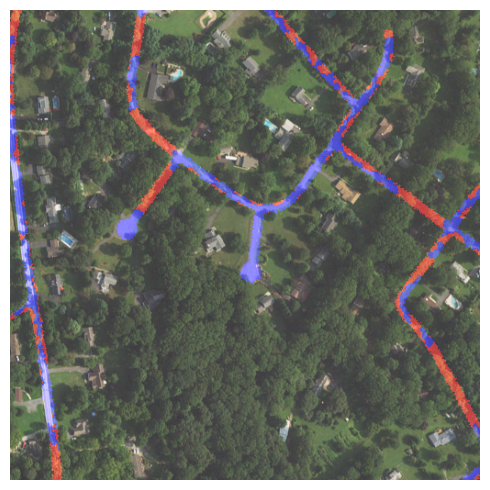}
    \includegraphics[width=0.25\textwidth]{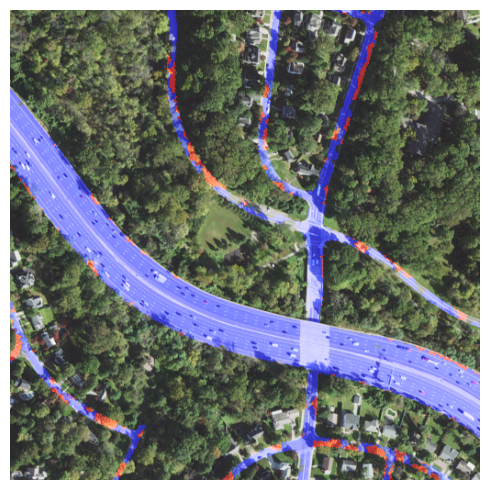}
    \includegraphics[width=0.25\textwidth]{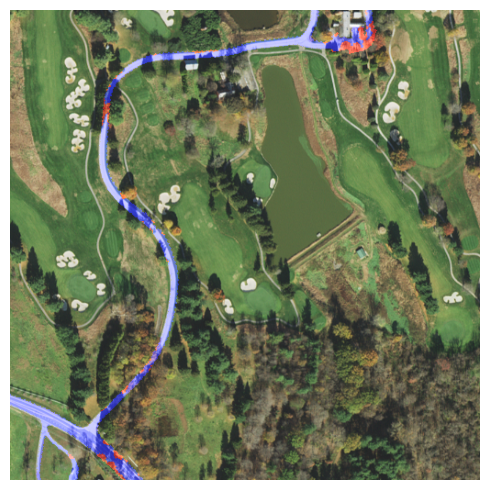} \\
    \includegraphics[width=0.25\textwidth]{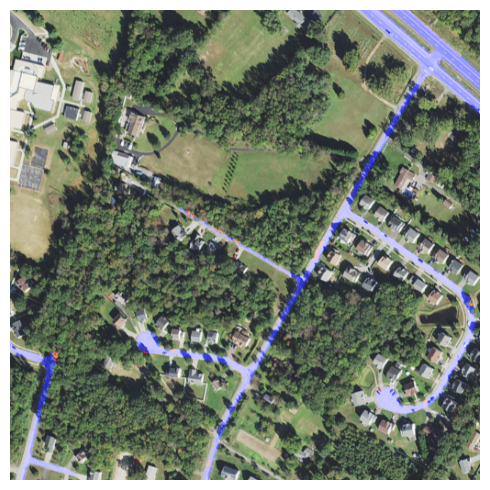}
    \includegraphics[width=0.25\textwidth]{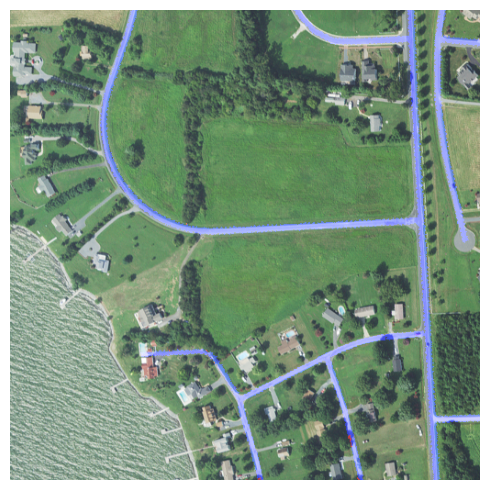}
    \includegraphics[width=0.25\textwidth]{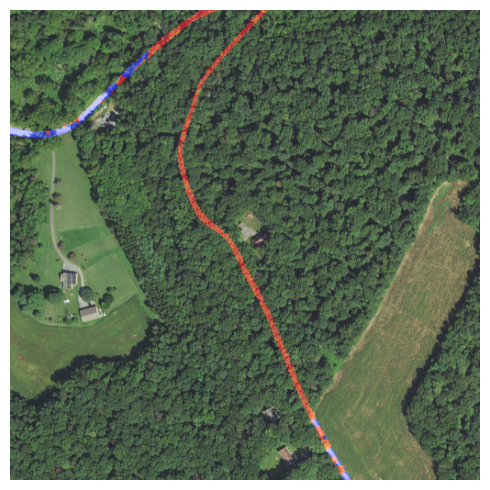}
    \caption{Example images and labels from the proposed \dataset dataset. Labels are shown over the corresponding NAIP aerial imagery with the ``Road'' class colored in \textcolor{blue}{blue} and the ``Tree Canopy over Road'' class in \textcolor{red}{red}.}
    \label{fig:dataset_examples}
\end{figure*}

\begin{figure}[t]
    \centering
    \includegraphics[width=0.45\linewidth]{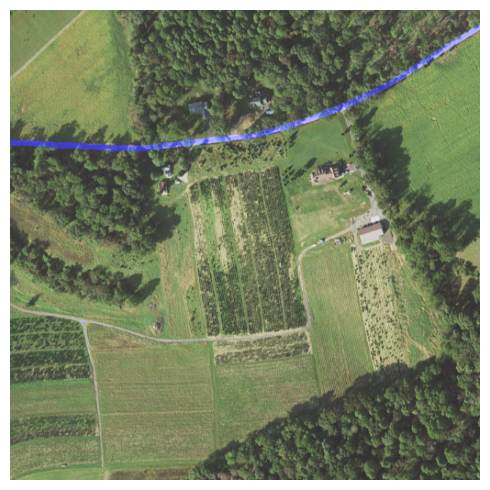}
    \includegraphics[width=0.45\linewidth]{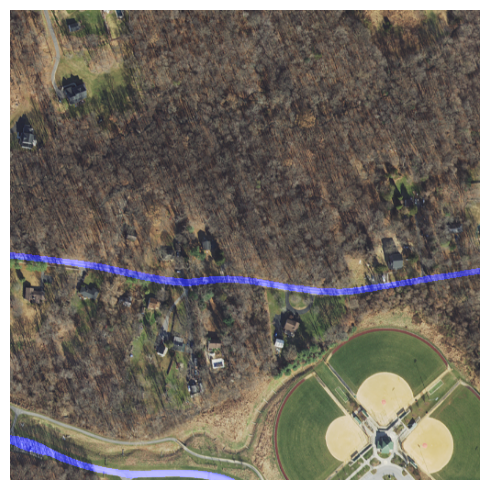}
    \includegraphics[width=0.45\linewidth]{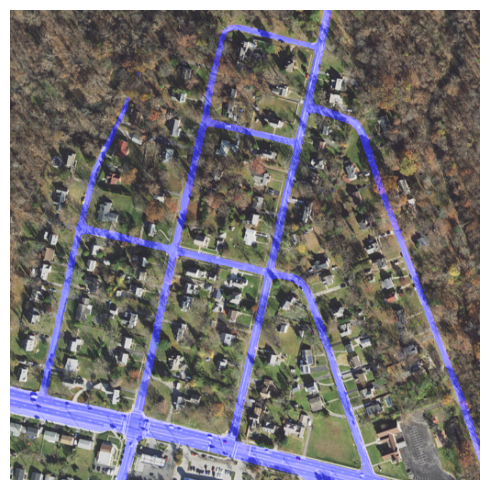}
    \includegraphics[width=0.45\linewidth]{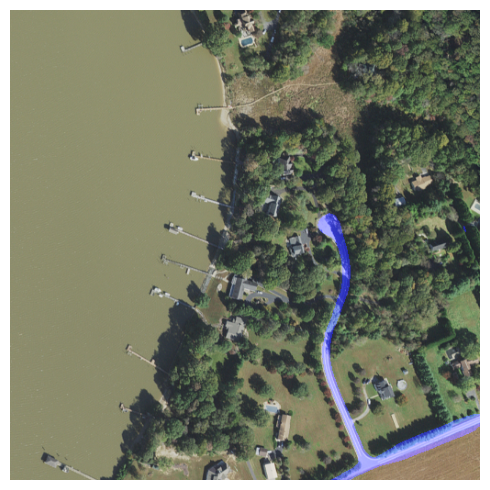}
    \caption{Example predictions (shown highlighted in \textcolor{blue}{blue}) from the U-Net ResNet-18 model on held out (test set) imagery. Note that the model is only trained to discriminate between ``road'' (including occluded and unoccluded roads) and ``background'' classes.}
    \label{fig:prediction_examples}
\end{figure}

\section{Methods}
\label{sec:methods}

\subsection{\dataset Dataset} \label{subsec:dataset}

We constructed the \dataset dataset to test the ability of semantic segmentation models to incorporate long-range dependencies in remotely sensed imagery. Specifically, we sample 30,000 \textit{patches} of aerial imagery from the US Department of Agriculture's (USDA) National Agricultural Imagery Program (NAIP) and label masks from the Chesapeake Bay Conservancy 2017/2018 land use land cover dataset~\cite{chesapeake2023land} over the US state of Maryland. Each \textit{patch} is $512 \times 512$ pixels at a $1\text{m}/\text{pixel}$ spatial resolution\footnote{We reproject and resample the NAIP imagery to the coordinate reference system and resolution of the land cover dataset.}, and contains 4 band (red, green, blue, near infrared) aerial imagery from 2018, and per-pixel land cover masks with ``background'', ``impervious roads'', and ``tree canopy over impervious roads'' classes. The land cover masks were created by the Chesapeake Conservancy~\cite{chesapeake2023land} using the same 2018 NAIP imagery as reference\footnote{For more information about this dataset, see the data dictionary and documentation at \url{https://doi.org/10.5066/P981GV1L}}. We sample patches to include relatively ``hard'' instances that contain over 10\% of ``tree canopy'' (we are interested in cases where the model will have the potential to confuse the ``tree canopy over road'' class with features present in the ``background'' class) and include all instances that contain the ``tree canopy over road'' class. In the final dataset 82.9\% of the patches contain the road class, 80.6\% contain the ``tree canopy over road class'' and 16.1\% only contain the ``background'' class. Although most patches contain instances of the ``road'' and ``tree canopy'' over road classes, the dataset is highly imbalanced -- 96.3\% of pixels are labeled as ``background'', 3.0\% of pixels are labeled as ``road'', and 0.7\% are labeled as ``tree canopy over road''. We observe that the distribution of distances from ``tree canopy over road'' labels to the nearest ``road'' label is highly skewed with a median of 4 pixels, a 95th percentile of 107 pixels and 99th percentile of 285 pixels. Finally, we divide the patches uniformly at random into fixed training, validation, and test splits with 80\%, 10\%, and 10\% proportions respectively. Figure \ref{fig:map} shows the spatial distribution of the patches and Figure \ref{fig:dataset_examples} shows several examples of patches from the dataset.

\subsection{Experimental setup}

We test the spatial reasoning of different semantic segmentation models by training the models with the ``road'' and ``tree canopy over road'' classes grouped (i.e. into an ``all roads'' class) but measuring performance on both subgroups. Our hypothesis is that models will need to use spatial context or reasoning in order to correctly classify pixels that are ``tree canopy over road'' as ``all roads'' by extrapolating from the existing road locations. For example, models that only depend on local color or texture features will always confuse ``tree canopy over road'' with ``background'' as ``tree canopy'' is part of the ``background'' class. 

We test the following modeling approaches:
\begin{description}
\item[Fully Convolutional Network (FCN)] This model is a simple baseline Convolutional Neural Network (CNN) architecture, composed of five convolutional layers with 128 3x3 kernels each followed by a ReLU (Rectified Linear Unit) nonlinearity. This structure is followed by a 1x1 convolutional layer used for classification. This model has a small receptive field~\cite{luo2016understanding} of only six pixels to limit its capacity to integrate substantial spatial context into per-pixel classifications.
\item[U-Net with ResNet-18 backbone] We use a standard U-Net architecture~\cite{ronneberger2015u} that incorporates a ResNet-18 backbone pretrained on ImageNet from the Segmentation Models PyTorch library~\cite{Iakubovskii:2019}. The U-Net model has a  receptive field of 286 pixels.
\item[U-Net with ResNet-50 backbone] This is the same approach as the above ``U-Net with ResNet-18 backbone'' but with a ResNet-50 backbone. This model has the same receptive field.
\item[DeepLabV3+ with ResNet-18 backbone] This is the same approach as the above, but with a DeepLabV3+ model~\cite{chen2018encoder} and a ResNet-18 backbone. The receptive field of this model is 978.
\item[DeepLabV3+ with ResNet-50 backbone] This is the same approach as above, but using a ResNet-50 backbone, with the same receptive field.
\end{description}

We train all models with a standard pixel-wise cross entropy loss and an AdamW optimizer~\cite{loshchilov2017decoupled} for 150 epochs. We use a cosine annealing learning rate schedule (without restarts) with a period of 100 epochs with an initial learning rate of 1e-3 and minimum of 1e-6.

We evaluate the models by measuring the precision and recall of the ``background'' and ``road'' classes in a standard way. We also measure the recall of the ``tree canopy over road'' class by counting ``road'' predictions as positives and ``background'' predictions as negatives. Additionally, we compute a distance weighted version of recall where each ``tree canopy over road'' label is weighted by its distance from the nearest ``road'' label. This will better measure how a model is performing on hard samples that require using larger spatial contexts to correctly classify each pixel.

\section{Results}
\label{sec:results}

\begin{table}[htb]
\centering
\resizebox{\columnwidth}{!}{%
\begin{tabular}{@{}ccccccc@{}}
\toprule
\multirow{2}{*}{\textbf{Model}} & \multicolumn{2}{c}{\textbf{Background}} & \multicolumn{2}{c}{\textbf{Road}} & \multicolumn{2}{c}{\textbf{\begin{tabular}[c]{@{}c@{}}Tree Canopy\\ over Road\end{tabular}}} \\ \cmidrule(lr){2-3}\cmidrule(lr){4-5}\cmidrule(lr){6-7}
 & \textbf{R} & \textbf{P} & \textbf{R} & \textbf{P} & \textbf{R} & \textbf{DWR} \\ \midrule
FCN & 99.4 & 98.3 & 64.1 & 71.5 & 23.4 & 10.7 \\
U-Net (ResNet-18) & 99.4 & 99.2 & 83.6 & 71.1 & 63.5 & \textbf{46.5} \\
U-Net (ResNet-50) & \textbf{99.5} & 99.2 & 84.0 & 71.8 & 63.5 & 45.7 \\
DeepLabV3+ (ResNet-18) & 99.4 & 99.1 & 82.6 & 71.9 & 58.9 & 41.3 \\
DeepLabV3+ (ResNet-50) & \textbf{99.5} & \textbf{99.3} & \textbf{84.8} & \textbf{72.0} & \textbf{63.9} & 46.1 \\ \bottomrule 
\end{tabular}
}
\caption{Test set performance of each method (P=precision, R=recall, DWR=distance weighted recall).}
\label{tab:results}
\end{table}

\begin{figure}
    \centering
    \includegraphics[width=1\linewidth]{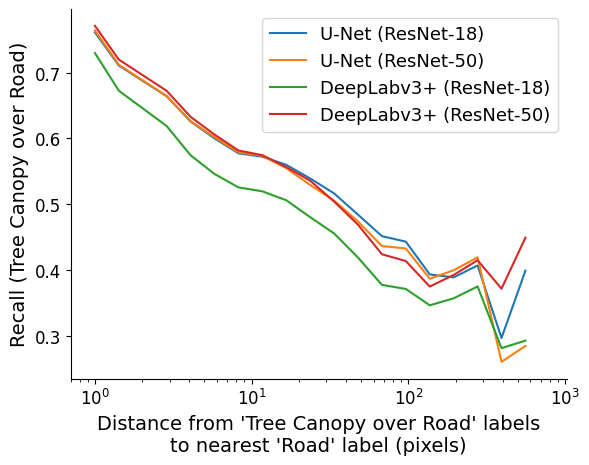}
    \caption{Recall of each model on the ``Tree Canopy over Road'' class shown as a function of distance from the nearest road class. Over all models, the performance on this class drops the farther away from the nearest road pixel the classification is made. For example, while the DeepLabv3+ ResNet-18 has $>70\%$ recall on ``tree canopy over road'' pixels that are adjacent to ``road'' pixels, the recall drops below $50\%$ for ``tree canopy over road'' pixels that are over 15 pixels away from a ``road'' pixel.}
    \label{fig:results_by_distance}
\end{figure}

Table \ref{tab:results} shows the test set performance of each model across the ``background'', ``road'', and ``tree canopy over road'' classes. Unsurprisingly, the baseline FCN model with its limited receptive field performs the worst -- even though half of the ``tree canopy over road'' labels are within 4 pixels of a road label (i.e. within the receptive field of the model) it is only able to achieve a recall of 23.4\% on the ``tree canopy over road'' class.
The other common semantic segmentation models all perform similarly, with the DeepLabV3+ ResNet-50 performing slightly better on all metrics but distance weighted recall (where the simplest model, a U-Net with ResNet-18 backbone performs best). We observe that the distance weighted recall over the ``tree canopy over road'' class is much lower than unweighted recall across all models ($\approx25\%$ for the common semantic segmentation models), meaning that the models are performing worse at correctly identifying ``tree canopy over road'' the farther away it is from a ``road'' class. We can see this explicitly in Figure \ref{fig:results_by_distance}. Here we bin ``tree canopy over road'' pixels according to their distance away from a ``road'' pixel and compute the recall of a model's predictions for each of the different distance bins. We see that recall for the ``tree canopy over road'' pixels that are closest to ``road'' pixels is relatively high (i.e. closer to that of the model's recall for the ``road'' class), however decreases with distance and is between 30\% and 40\% for the pixels that are farthest away from roads. This shows the models are strongly biased towards using local information in their classifications. 

\section{Conclusion}
\label{sec:conclusion}
We propose the \dataset dataset for studying how semantic segmentation models incorporate spatial context into their predictions and perform benchmark experiments with common models. We find that common semantic segmentation models are not able to well model a ``tree canopy over road'' class, where spatial context is needed to make a correct classification. The performance of these models further decreases with the distance away from relevant context (i.e. distance to the nearest road). Future work can focus on an in-depth study and comparison of CNN variants, transformers, and sequence-to-sequence- networks (for example, that measures effective receptive fields) or how to improve spatial reasoning in geospatial ML models.

\bibliographystyle{IEEEbib}
\bibliography{refs}

\end{document}